\documentclass[lettersize,journal]{IEEEtran}
\usepackage{amsmath,amsfonts}
\usepackage{amssymb}
\usepackage{algorithmic}
\usepackage{algorithm}
\usepackage{array}
\usepackage[caption=false,font=normalsize,labelfont=sf,textfont=sf]{subfig}
\usepackage{textcomp}
\usepackage{stfloats}
\usepackage{url}
\usepackage{verbatim}
\usepackage{graphicx}
\usepackage{multirow}
\usepackage{textalpha}
\usepackage{textgreek}
\usepackage{cite}
\usepackage{xcolor}
\newcolumntype{P}[1]{>{\centering\arraybackslash}p{#1}}
\newcolumntype{M}[1]{>{\centering\arraybackslash}m{#1}}

\usepackage{hyperref}
\hypersetup{
    colorlinks=true,
    linkcolor=blue,
    filecolor=magenta,      
    urlcolor=blue,
    citecolor=black,
    pdftitle={YOLO3D Final Report},
    pdfpagemode=FullScreen,
    }

\begin{document}

\title{OriCon3D: Effective 3D Object Detection using
Orientation and Confidence}

\author{Dhyey Manish Rajani$^{1}$, Rahul Kashyap Swayampakula$^{1*}$, Surya Pratap Singh$^{1*}$}

\maketitle


\begin{abstract} 
In this paper, we propose an advanced methodology for the detection of 3D objects and precise estimation of their spatial positions from a single image. Unlike conventional frameworks that rely solely on center-point and dimension predictions, our research leverages a deep convolutional neural network-based 3D object weighted orientation regression paradigm. These estimates are then seamlessly integrated with geometric constraints obtained from a 2D bounding box, resulting in derivation of a comprehensive 3D bounding box. Our novel network design encompasses two key outputs. The first output involves the estimation of 3D object orientation through the utilization of a discrete-continuous loss function. Simultaneously, the second output predicts objectivity-based confidence scores with minimal variance. Additionally, we also introduce enhancements to our methodology through the incorporation of lightweight residual feature extractors. By combining the derived estimates with the geometric constraints inherent in the 2D bounding box, our approach significantly improves the accuracy of 3D object pose determination, surpassing baseline methodologies. Our method is rigorously evaluated on the KITTI 3D object detection benchmark, demonstrating superior performance.
\end{abstract}


\section{Introduction}

\footnotetext[1]{The authors are with the Department of Robotics, University of Michigan, 48109 Ann Arbor, USA}
\def\thefootnote{*}\footnotetext{Equal contribution; these authors can swap ordering as per their need}\def\thefootnote{\arabic{footnote}}
\footnotetext[2]{Corresponding e-mail: drajani@umich.edu; Other authors email: rahulswa@umich.edu; suryasin@umich.edu} 

\IEEEPARstart{M}{} onocular 3D object detection is extensively utilized in autonomous driving systems, specifically in advanced driver-assistance systems (ADAS), to comprehend the external environment. In this context, the sole sensory input for determining an object's location, dimensions, and orientation comes from images captured by a single camera. Despite the presence of multiple cameras in most vehicles, the necessity of monocular 3D object detection persists, as a significant portion of the surroundings is visible only to a single camera. The accurate detection of objects in 3D space, including cars, pedestrians, cyclists, etc., is fundamental for vehicles to execute subsequent driving tasks such as collision avoidance and adaptive cruise control. While 2D detection algorithms have been successful in handling variations in viewpoint and clutter, accurately detecting 3D objects remains a difficult problem. Previous attempts to combine pose estimation with object detection have primarily focused on estimating viewpoint by dividing it into discrete categories that can be trained, based on the fact that an object's appearance changes with viewpoint\cite{xiang2015data}. In situations where full 3D pose estimation is not possible, alternative approaches involve sampling and scoring hypotheses using contextual and semantic cues\cite{chen2016monocular}.
\begin{figure}[h]
\captionsetup{justification=centering}
\includegraphics[width=8cm, height=7cm]{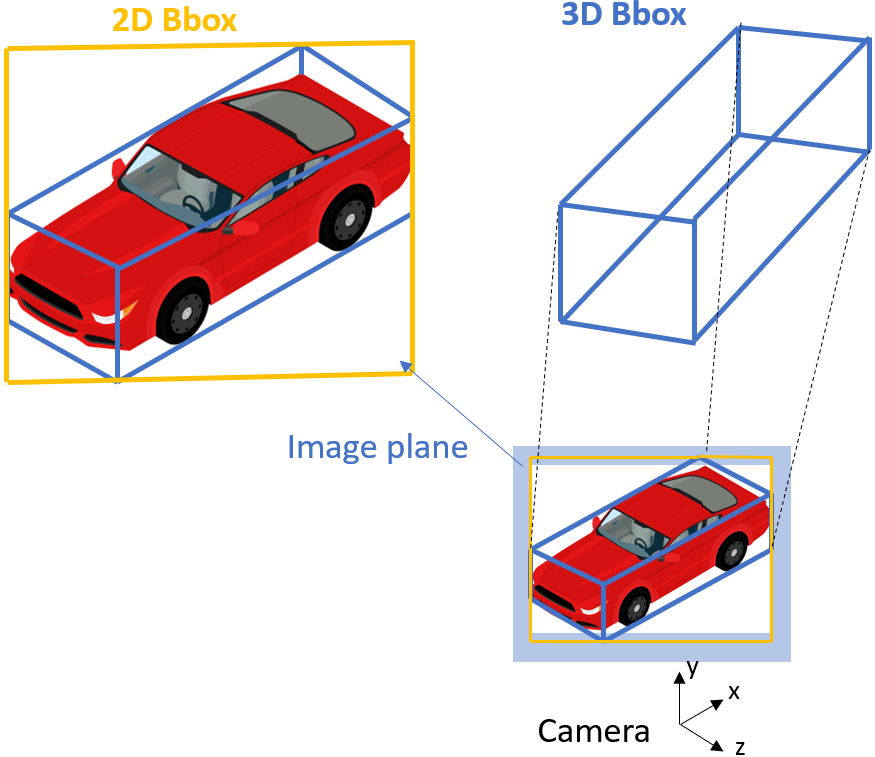}
\caption{3D bounding box projection into 2D detection window for enhanced orientation and confidence estimation}
\label{fig:cover}  
\end{figure}
In our work, we present a two-stage approach to accurately estimate the 3D bounding box pose, (R, T) $\epsilon$ SE(3), without directly regressing over the bounding box dimensions from a single image. First, we utilize a pre-trained 2D bounding box estimator i.e. YOLO (You-only-look-once)-v3\cite{redmon2018yolov3} on the input image to constrain the 3D bounding box's projection, ensuring it fits within the 2D detection window and yielding the 3D bounding box's orientation. Despite its simplicity, this method builds upon several important insights. One such insight is that not directly regressing to vehicle dimensions is effective, given the low variance in bounding box dimensions across each subcategory of the KITTI 3D Object Detection dataset\cite{geiger2012we}. Additionally, we employ the MultiBin discrete-continuous formulation introduced in \cite{mousavian20173d} for orientation regression, as it has been shown to outperform traditional L2 loss. This orientation estimate is then combined with geometric object constraints to obtain the 3D bounding box's center coordinates. To enhance the applicability of our method on mobile devices, we conduct an ablation study with lightweight backbone feature extractors, namely MobileNet-v2\cite{howard2017mobilenets} and EfficientNet-v2\cite{tan2021efficientnetv2}. After the separate integration of these feature extractors, all the models are trained individually from scratch on the KITTI 3D Object Detection dataset. Notably, these backbone networks not only improved inference time but also yielded state-of-the-art performance across various metrics.

We assess the performance of our method on both the KITTI 3D Object Detection validation dataset \cite{chen20153d} and the KITTI 3D Object Detection test dataset\cite{geiger2012we}, simultaneously comparing our estimated 3D boxes with the results from other contemporary and state-of-the-art approaches, mentioned in Section \ref{II}. Although the official benchmark for 3D bounding box estimation solely evaluates the 3D box orientation estimate, we go beyond by incorporating three additional performance metrics to gauge accuracy: the distance to the box's center, the distance to the center of the nearest bounding box face, and the overall overlap between the bounding box and the ground truth box, measured using the 3D Intersection over Union (3D IoU) score. Our results demonstrate that, with sufficient training data, our method outperforms state-of-the-art algorithms across all these 3D metrics.

\section{Related Works}
\label{II}
Earlier, researchers addressed the challenge of determining the position and orientation of objects in 3D space exclusively using LiDARs\cite{shi2019pointrcnn}. However, LiDARs are both expensive and inefficient, particularly in adverse weather conditions\cite{tang2020center3d} and glassy environments. Consequently, numerous studies have redirected their attention to 3D object detection based on monocular cameras—a more direct approach, albeit susceptible to scale/depth ambiguity\cite{tang2020center3d}. To address this challenge, one method involves employing a mathematical technique known as the perspective n-point problem (PnP). This technique entails matching 2D points in the image with their corresponding 3D points in a model of the object. Various approaches exist, including closed-form or iterative solutions \cite{lepetit2009ep}, as well as constructing 3D models of the object and finding the best match between the model and the image \cite{ferrari2006simultaneous}\cite{rothganger20063d}. Earlier methods\cite{chen2016monocular}\cite{rath2020boosting} utilized hand-crafted features, while recent approaches leverage deep learning. Some variations include modifying architectures \cite{liu2022learning}\cite{tang2020center3d} or adjusting loss functions\cite{brazil2019m3d}\cite{chen2020monopair}. Specific methods consider scale into consideration\cite{lu2021geometry}, while others integrate depth into convolution\cite{brazil2019m3d}\cite{ding2020learning} or factor in confidence\cite{brazil2020kinematic}\cite{kumar2020luvli}. Recent developments involve in-network ensembles for deterministic\cite{zhang2021objects} or probabilistic depth predictions\cite{lu2021geometry}. A subset of these methods incorporates temporal cues\cite{brazil2020kinematic}, non-maximum suppression (NMS)\cite{kumar2021groomed}, or LiDAR\cite{reading2021categorical} during the training process. Another noteworthy approach, known as Pseudo-LiDAR\cite{park2021pseudo}\cite{simonelli2021we}, first estimates depth and subsequently employs a point cloud-based 3D object detector.

New datasets \cite{geiger2012we}\cite{matzen2013nyc3dcars}\cite{xiang2016objectnet3d}\cite{xiang2014beyond} have made it possible to extend 3D pose estimation to include entire object categories, dealing with changes in appearance due to different poses and variations within the category\cite{kar2015category}\cite{pepik20153d}. To tackle this problem, some researchers \cite{pepik2012teaching}\cite{xiang2014beyond} have used a framework called discriminative part-based models (DPMs) within an object detection framework. This approach considers each mixture component to represent a different azimuth section and solves the problem of pose estimation as a structured prediction problem. However, these methods only predict a subset of Euler angles with respect to the object's canonical frame and do not estimate the object's position or dimensions.

An alternative approach to 3D pose estimation is to make use of existing 3D shape models of objects. This involves sampling hypotheses about the object's viewpoint, position, and size, and then comparing rendered 3D models of the object to the detection window in the image using HOG features\cite{mottaghi2015coarse}. Researchers have also explored using CAD models to estimate object poses\cite{zhu2014single}, but this has only been successful in simpler settings with less challenging detection problems. In more difficult scenarios, such as those with significant occlusion, researchers have developed methods that use dictionaries of 3D voxel patterns learned from CAD models to characterize both the object's shape and common occlusion patterns\cite{xiang2015data}, refining the object's pose by estimating correspondences between the projected 3D model and image contours. These methods have been evaluated on datasets such as PASCAL3D+ and in tabletop settings with limited clutter or scale variations.

Several recent methods have been developed to detect 3D bounding boxes for driving scenarios, and our method is most closely related to them. Xiang et al. used 3D voxel patterns to cluster the possible object poses into viewpoint-dependent subcategories, capturing shape, viewpoint, and occlusion patterns. They used 3D CAD models to learn the pattern dictionaries and classified the subcategories discriminatively using deep CNNs. Chen et al. tackled the problem by sampling 3D boxes in the physical world and scoring them based on high-level contextual, shape, and category-specific features, assuming the flat ground plane constraint. However, all of these approaches require complicated preprocessing, such as segmentation or 3D shape repositories, and may not be suitable for robots with limited computational resources.

\section{Datasets}
KITTI is a widely used dataset for mobile robots and autonomous driving, encompassing several hours of recorded traffic scenarios captured multimodally with sensors such as high-definition RGB, grayscale stereo cameras, and 3D laser-scanners. The dataset features image sequences with a resolution of 1242×375, taken at 10 FPS (frames per second) from a VW wagon traveling in urban environments. The benchmark for 3D object detection\cite{geiger2015kitti} comprises 7,481 training and 7,518 test images\cite{geiger2013vision}, each accompanied by associated point clouds. To facilitate training, we partition the set of 7,481 training images into two subsets: 3,712 for training and 3,769 for validation\cite{chen20153d}. Object annotations in the dataset include 2D bounding boxes with six class labels: cars, pedestrians, cyclists, trams, trucks, and vans. The 3D Object Detection dataset is further categorized into three classes, as outlined in Table \ref{table:dataset}, and these categories were used during the evaluation process.


\begin{table}[htb]
    \centering
    \caption{KITTI 3D Object detection dataset difficulty levels}
    \begin{tabular}{|M{1.6cm}|M{1.8cm}|M{1.8cm}|M{1.8cm}|}
        \hline
        Difficulty level & Min. bounding box height & Max. truncation & Max. occlusion level \\
        \hline
        Easy & 40 Px & 15{\%} & Fully visible  \\
        \hline
        Moderate & 25 Px & 30{\%} & Partly occluded \\
        \hline
        Hard & 25 Px & 50{\%} & Difficult to see \\
        \hline
        
    \end{tabular}
    \label{table:dataset}
\end{table}

\section{Methodology}
\label{IV}

Our architecture as shown in Fig. II, consists of two main parts: a deep convolutional neural network (CNN) and a geometric reasoning module. The CNN is responsible for feature extraction and produces a feature map that encodes spatial \& semantic information about the objects in the image. The geometric reasoning module takes the feature map as input and generates 3D bounding box proposals by combining the CNN output with geometric constraints based on the object's projected 2D bounding box.

\subsection{Feature extraction}

We thoroughly assessed the suitability of three distinct backbone networks, each varying in depth, for the intricate task of 3D object detection from a single 2D image. Among the featured extractors, the VGG-19 network emerged as a prominent choice—a widely recognized CNN architecture for image classification. Boasting 19 layers, including 16 convolutional layers and 3 fully connected layers\cite{simonyan2014very}, VGG-19 employs sequential convolutional layers with 3x3 filters in the initial 13 layers and 1x1 filters in the subsequent 6. Max-pooling is applied after every two convolutional layers. The input, a 224x224x3 RGB image, undergoes transformation into a 7x7 feature vector with 512 channels. In our pipeline, this vector serves as input for a separate network dedicated to 3D bounding box estimation.

\begin{figure*}
\captionsetup{justification=centering}
\centering
\includegraphics[width=18cm, height=5cm]{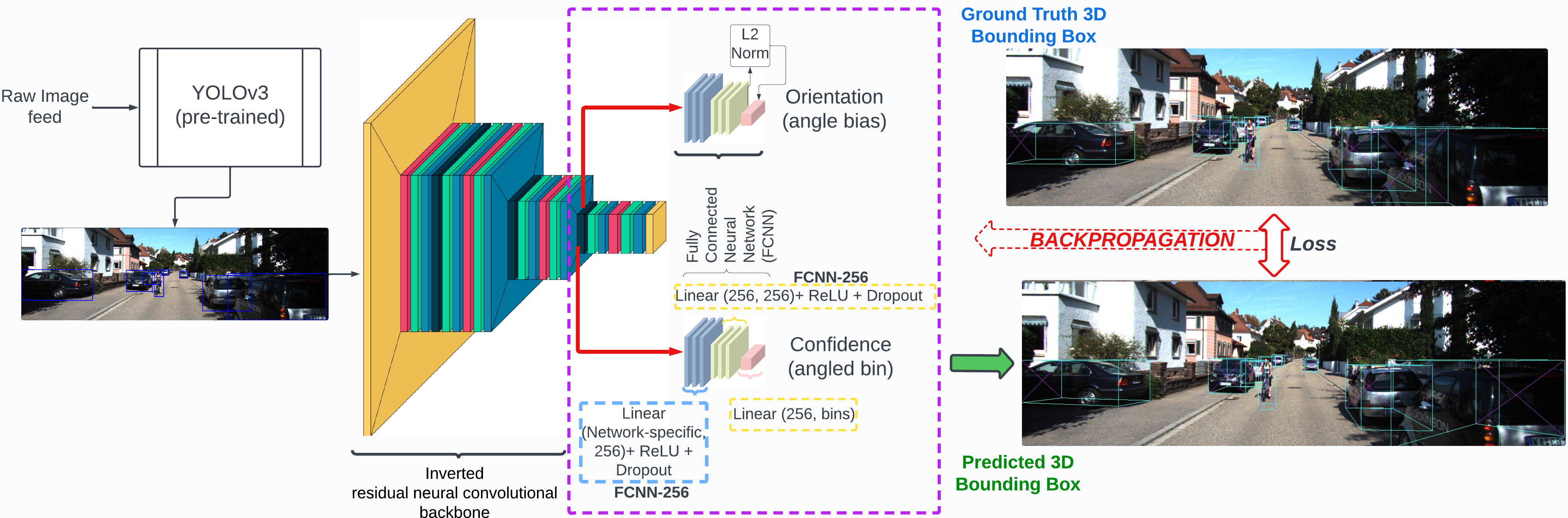}
\caption{OriCon3D: Our customized multi-bin multitask learning-based orientation and confidence estimator}
\label{fig:architecture}  
\end{figure*}

Another architecture we've examined is MobileNet-v2, purposefully designed for efficient computation on mobile and embedded devices\cite{howard2017mobilenets}. Our adapted version takes a 224x224x3 RGB image as input, yielding a 1280x7x7 output. Employing depthwise separable convolutional layers, MobileNet-v2 maintains high accuracy while minimizing computational cost, rendering it ideal for deployment on mobile devices. The architecture initiates with a standard convolutional layer with a stride of 2, reducing the spatial dimensionality of the input feature map. Subsequently, a series of depthwise separable convolutional layers follows, incorporating batch normalization and ReLU activation. The network achieves optimal performance with its "bottleneck" layers, leading to efficient feature representations. These features are then passed to downstream regression layers for 3D bounding box estimation.

In our exploration of backbone networks, we've delved into the EfficientNet-v2 family, opting for the EfficientNetv2-S—the most lightweight member tailored for mobile and edge devices with constrained computational resources\cite{tan2021efficientnetv2}. The network's input is a 224x224x3 RGB image, producing an output of 512x7x7. The architecture features stem convolutional layers, various convolutional blocks, and a top section dedicated to classification or regression tasks. Employing techniques like efficient channel gating, weight standardization, and compound scaling, EfficientNet-v2 models excel in achieving heightened performance while adhering to computational efficiency. The network's ability to strike a balance between depth, width, and resolution showcases its versatility in addressing diverse computer vision tasks, making it particularly adept for various deployment scenarios.

\subsection{Geometric reasoning}
The geometric reasoning module consists of three main components: a 2D bounding box estimator, a 3D bounding box estimator, and a refinement module. The 2D bounding box estimator uses a deep multi-layer convolutional perceptron to predict the 2D bounding box coordinates of the object in the image. The 3D bounding box estimator takes the 2D bounding box coordinates as input and predicts the 3D bounding box dimensions and orientation. Finally, the refinement module refines the 3D bounding box proposals by minimizing the geometric error between the predicted and ground-truth 3D bounding boxes.
    \subsubsection{Generalized 3D Bounding Box Estimation}
    To make use of the success of 2D object detection in estimating 3D bounding boxes, we rely on the fact that the projection of a 3D bounding box onto a 2D image should fit tightly within its 2D detection window \cite{mousavian20173d}. We assume that YOLO-based 2D object detector has been trained to produce boxes that correspond to the bounding box of the projected 3D box. The 3D bounding box is defined by its center $T = [t_{x}, t_{y}, t_{z}]$, dimensions $D = [d_{x}, d_{y}, d_{z}]$, and orientation R($\theta$, φ, α), which are determined by the azimuth, elevation, and roll angles. Given the position and orientation of the object in the camera coordinate system $(R, T) \epsilon SE(3)$ and the camera intrinsics matrix K, we can project a 3D point $Xo = [X, Y, Z, 1]^T$ in the object's coordinate system onto the image plane as $x = [x, y, 1]^T$ is: 
    \begin{equation}
    x = K[R\; T]Xo
    \end{equation}
    If we assume that the center of the 3D bounding box corresponds to the origin of the object coordinate frame, and we know the dimensions of the object (D), then the coordinates of the 3D bounding box vertices can be described by simple equations, such as $X1 = [d_{x}/2, d_{y}/2, d_{z}/2]^T$, $X2 = [-d_{x}/2, d_{y}/2, d_{z}/2]^T$, and so on, up to $X8 = [-d_{x}/2, -d_{y}/2, -d_{z}/2]^T$. To ensure that the 3D bounding box fits tightly into the 2D detection window, we require that each side of the 2D bounding box is touched by the projection of at least one of the 3D box corners. For instance, if the projection of one 3D corner, $X0 = [d_{x}/2, -d_{y}/2, d_{z}/2]^T$, touches the left side of the 2D bounding box with coordinate $x_{min}$, this results in an equation that relates the two, as follows:
    \begin{equation}
    x_{min} = \begin{pmatrix}K[R\; T]\begin{bmatrix} d_{x}/2 \\ -d_{y}/2 \\ d_{z}/2 \end{bmatrix} \end{pmatrix}_{x}
    \end{equation}
    In the above equation, $(.)_{x}$ refers to the x coordinate of the projected point. We can derive similar equations for the other parameters of the 2D bounding box, such as $x_{max}$, $y_{min}$, and $y_{max}$. Overall, the sides of the 2D bounding box provide four constraints on the 3D bounding box. However, to fully determine the nine degrees of freedom (DoF) of the 3D box, which include three for translation, three for rotation, and three for box dimensions, generalized conventional methods constrain the 3D box to estimate additional geometric properties from its visual appearance. These properties are strongly related to the visual appearance and provide additional constrained parameters on the final 3D box.

   Along with the orientation parameters ($\theta$, φ, α), instead of embracing translation box methodologies, prevalent approaches tend to lean towards dimension regression, presuming minimal variation in object size. However, a notable drawback surfaces due to the utilization of dimension regression, introducing a high number of parameters in the neural network and potentially extending training times. Moreover, these orientation parameters are intertwined with the process of ascertaining translation parameters of the 3D bounding box. Conventional methods often rely on dimensions and orientations obtained through CNN regression and the 2D detection box. While this approach proves effective, an associated disadvantage emerges. The exhaustive exploration of all possible correspondences between 2D box sides and 3D box corners, initially resulting in 4096 configurations, demands computational resources. Although this is later optimized to 64 configurations by assuming an upright object and near-zero relative roll, the exhaustive search presents challenges, particularly in scenarios with resource constraints.

    \subsubsection{MultiBin Orientation Estimation}
\begin{figure}[hbtp]
    \centering
    \includegraphics[width= 6cm, height=6cm]
    {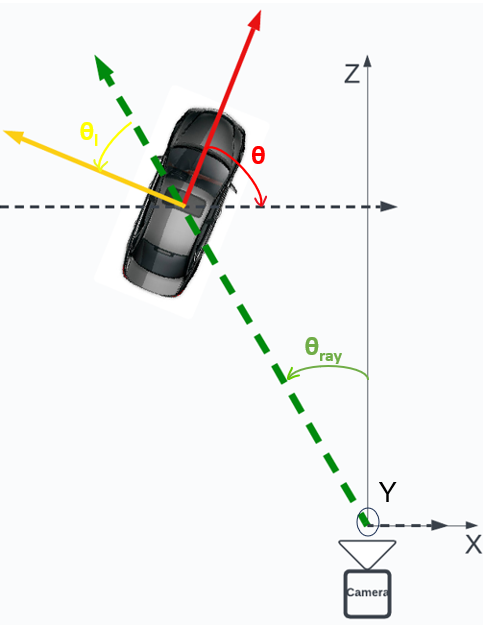}
    \caption{The local $(\theta_{l})$ egocentric, global allocentric orientation $(\theta)$ of the vehicle, and ray angle $(\theta_{ray})$ w.r.t camera centre is shown. Vehicle's heading is shown by red arrow and green arrow is the centre ray i.e. between origin and vehicle body center. Therefore, vehicle orientation ($\theta$) = $\theta_{ray}$ + $\theta_{l}$. The network is trained to regressively estimate $\theta_{l}$.}
    \label{fig:car}
    \end{figure}

    Estimating an object's global orientation in camera frame from detection window localized region requires the region's crop position within the image plane. In the case of a car moving in a straight line, even though global orientation of the car doesn't change, its local orientation w.r.t. the ray through crop center does change, resulting in appearance changes of the cropped image. The rotation of the car is only parameterized by azimuth (yaw) angle, as seen in Figure \ref{fig:car}. Hence, we estimate the local orientation angle of the object w.r.t. the ray passing through the crop center. In Figure \ref{fig:car}, the local orientation angle \& ray angle change in a way resulting in constant global orientation of the car. During inference, we use the ray direction and estimated local orientation to compute the global object orientation.

    In multi-task 3d object detection there are five branches used in \cite{fang20193d}, where the each branch corresponds to computing dimension residual, angle residual, confidence of each bin, viewpoint classification, and the center projection of bottom face (cbf) respectively. In\cite{mousavian20173d}, there are three branches each corresponding to dimensions, angle bias, and confidence probability. Whereas in our architecture, shown in Fig. \ref{fig:architecture}, we have used a multibin multi-task learning considering only two branches: angle residual and confidence of each bin.

    Leveraging the MultiBin \cite{mousavian20173d} architecture approach in our method, the orientation angle is divided into n overlapping discrete bins. For each bin, the CNN network estimates a confidence probability $(c_{i})$ (that output angle falls within that bin) \& residual rotation (to obtain the output angle from orientation of the center ray of that bin). The residual rotation is represented by sine and cosine of the angle, resulting in 3 outputs per bin: ($c_{i}$, cos($\Delta$$\theta_{i}$), sin($\Delta$$\theta_{i}$). The cosine and sine values are obtained by applying an L2 normalization layer to a 2-dimensional input. The total loss for the MultiBin orientation is thus calculated based on these outputs.

    \begin{equation}
    L_{θ} = L_{conf} + w \times L_{loc} 
    \end{equation}

    The confidence loss $(L_{conf})$ is calculated using the softmax loss of all bins. The localization loss $(L_{loc})$ minimizes the difference between the estimated and ground truth angle in each of the bins that cover the ground truth angle, with adjacent bins having overlapping coverage. The localization loss seeks to minimize the difference between the ground truth and all the bins that cover that value, which is equivalent to maximizing cosine distance, as shown in \cite{mousavian20173d}. The $L_{loc}$ is computed as follows:
    \begin{equation}
    L_{loc} = \frac{-1}{n_{\theta*}} \times \sum cos(\theta^* - c_{i} - \Delta\theta_{i})
    \end{equation}
    The equation refers to the number of bins that cover the ground truth angle $(n_{\theta
    *})$, the center angle of bin i ($c_{i}$), and the change in angle that needs to be applied to the center of bin i ($\Delta$$\theta_{i}$).

    During the inference stage, the bin with the highest confidence value is chosen, and the final output is calculated by applying the estimated $\Delta$$\theta$ of that bin to the center of that bin. The MultiBin architecture consists of two branches - one for computing the confidences $c_{i}$ and other for computing the cosine and sine of $\Delta\theta$. Therefore, the model needs to estimate 3n parameters for n bins.

\begin{figure*}
\centering
\captionsetup{justification=centering}
\includegraphics[width=18cm, height=7cm]{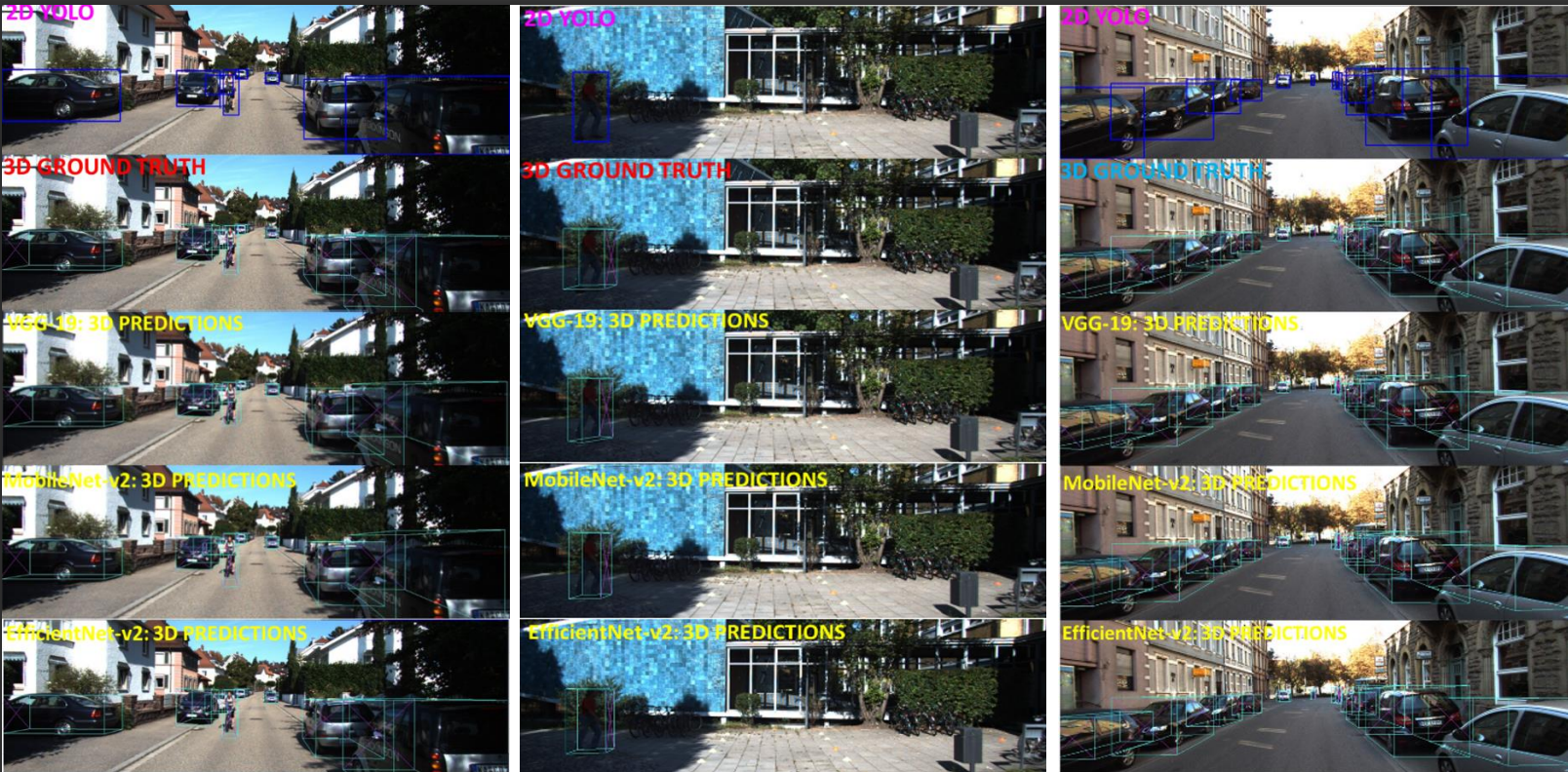}
\caption{Qualitative results: left column—cyclists and cars; middle column—pedestrians only; right column—cars only}
\label{fig:qualitative}  
\end{figure*}

In our algorithm, we first replace the head of multibin algorithm \cite{mousavian20173d} with a suitable light-weight extractor then using just the sine \& cosine of the orientation angle along with the confidence score, we formulate an equally weighted equation between them as our loss function, instead of using centers  or dimensions of bounding box as done in \cite{mousavian20173d}. According to us using orientation angle (sine + cosine), along with confidence score, can be an effective approach.

First, by using just orientation angle with confidence score, the model can focus on learning the essential information needed for 3D bounding box estimation, while ignoring other potentially less relevant information, hence reducing risk of over-fitting, which is especially important in our case of limited training data. Second, when using the raw orientation angle, it is difficult for the model to distinguish between angles that are 360 \textdegree apart. By using the sine and cosine, the model can represent the orientation information in a periodic \& continuous feature space, hence improving accuracy \& prediction stability. Third, the confidence score provides a measure of the model's certainty in its predictions. By using the confidence score as a weight, the model can assign more weight to confident predictions and less weight to uncertain ones. This can help improve the Average Precision (AP) of the 3D bounding box estimation.

After thorough experimentation and close observation we find that through this approach the dimensions and centre point can eventually be ascertained if the orientation of the bounding box is correct. Regressing for them on top of angle and confidence also increase training efficiency in large datasets, but considering the limited dataset \& computing resources we proved in the results section that our simple combination of inverted residual feature extraction \& multibin architecture proves to be more robust (or comparable in some cases) and memory efficient than the existing frameworks. Within the KITTI dataset, the dimensions for cars and cyclists exhibit only minor variations, typically within a few centimeters. When conducting regression heavily weighted along bounding box dimensions, false positive bounding box detections may arise. In contrast to employing our discrete-continuous loss-based MultiBin loss, the conventional approach involves directly using a dimension-based L2 loss function leading to lower-generalizability and increased time complexity.

\begin{table*}[ht!b]
    \centering
    \caption{ Backbone networks: Comparison of the AOS, AP, OS, IOU 3D on the KITTI Val dataset\cite{chen20153d} }
    \begin{tabular}{ |M{3.4cm}|M{0.8cm}M{0.8cm}M{0.8cm}M{0.8cm}|M{0.8cm}M{0.8cm}M{0.8cm}M{0.8cm}|M{0.8cm}M{0.8cm}M{0.8cm}M{0.8cm}|}
        \hline
        \multicolumn{1}{|c|}{} & \multicolumn{4}{c|}{Easy} & \multicolumn{4}{c|}{Moderate}
        & \multicolumn{4}{c|}{Hard} \\
        Class (with OriCon3D backbone) & AOS(\%) & AP(\%) & OS & IOU 3D & AOS(\%) & AP(\%) & OS & IOU 3D & AOS(\%) & AP(\%) & OS & IOU 3D\\
        \hline
        \multirow{1}{*}{Cars (with VGG-19)} & 96.06 & 30.68 & 0.997 & 0.861 & 89.70 & 24.79 & 0.995 & 0.744 & 82.38 & 22.91 & 0.989 & 0.702\\
        \multirow{1}{*}{Cars (with MobileNet)} & 96.00 & 30.77 & \textbf{0.998} & \textbf{0.899} & \textbf{89.77} & 24.85 & \textbf{0.996} & \textbf{0.782} & 82.38 & 23.10 & \textbf{0.989} & 0.713\\
        \multirow{1}{*}{Cars (with EfficientNet)} & \textbf{96.11} & \textbf{30.82} & 0.995 & 0.860 & 89.73 & \textbf{25.39} & 0.972 & 0.740 & \textbf{82.55} & \textbf{23.45} & 0.967 & \textbf{0.732} \\
        \hline
        \multirow{1}{*}{Pedestrian (with VGG-19)} & 79.93 & 19.72 & 0.926 & 0.778 & 72.91 & 16.00 & 0.911 & 0.626 & 66.55 & 14.51 & 0.892 & 0.629\\
        \multirow{1}{*}{Pedestrian (with MobileNet)} & \textbf{81.81} & 19.97 & \textbf{0.937} & \textbf{0.803} & \textbf{74.59} & 16.28 & \textbf{0.921} & \textbf{0.667} & 68.49 & 14.55 & 0.906 & 0.621\\
        \multirow{1}{*}{Pedestrian (with EfficientNet)} & 80.81 & \textbf{19.99} & 0.927 & 0.778 & 73.39 & \textbf{16.43} & 0.900 & 0.628 & \textbf{70.04} & \textbf{14.78} & \textbf{0.939} & \textbf{0.630}\\
         \hline
        \multirow{1}{*}{Cyclist (with VGG-19)} & 86.22 & 16.20 & 0.979 & 0.759 & 80.20 & 10.83 & 0.966 & 0.626 & 72.59 & 9.22 & 0.940 & 0.610\\
        \multirow{1}{*}{Cyclist (with MobileNet)} & 86.54 & 16.22 & 0.977 & 0.757 & \textbf{81.34} & 11.08 & \textbf{0.969} & \textbf{0.658} & 73.92 & 9.49 & \textbf{0.946} & 0.620\\
        \multirow{1}{*}{Cyclist (with EfficientNet)} & \textbf{86.98} & \textbf{16.35} & \textbf{0.987} & \textbf{0.823} & 80.46 & \textbf{11.34} & 0.948 & 0.633 & \textbf{74.25} & \textbf{9.75} & 0.939 & \textbf{0.680}\\
         \hline        
    \end{tabular}
    \label{table:backbone}
\end{table*}



   \begin{table}[htb]
    \caption{ Results on KITTI Val Cars\cite{chen20153d} at $IoU_{3D}$ $\geq$0.7}
    \begin{tabular}{ |M{3cm}|M{0.8cm}|M{0.8cm}M{0.8cm}M{0.8cm}|}
        \hline
        \multicolumn{1}{|c|}{} & \multicolumn{1}{c|}{Extra} & \multicolumn{3}{c|}{$AP_{3D|R_{40}}$(\%)}\\
        Method & training & Easy & Moderate & Hard\\
        \hline
        \multirow{1}{*}{DDMP-3D\cite{wang2021depth}} & Depth & 28.12 & 20.39 & 16.34\\
        \multirow{1}{*}{PCT\cite{wang2021progressive}} & Depth & 38.39 & 27.53 & 24.44\\
        \multirow{1}{*}{MonoDistill\cite{chong2022monodistill}} & Depth & 24.31 & 18.47 & 15.76\\
        \multirow{1}{*}{CaDDN\cite{reading2021categorical}} & LiDAR & 23.57 & 16.31 & 13.84\\
        \multirow{1}{*}{PatchNet-C\cite{simonelli2021we}} & LiDAR & 24.51 & 17.03 & 13.25\\
        \multirow{1}{*}{Kinematic\cite{brazil2020kinematic}} & Video & 19.76 & 14.10 & 10.47\\
        \hline
        \multirow{1}{*}{MonoRCNN\cite{shi2021geometry}} & - & 16.61 & 13.19 & 10.65\\
        \multirow{1}{*}{MonoDLE\cite{ma2021delving}} & - & 17.45 & 13.66 & 11.68\\
        \multirow{1}{*}{GrooMeD-NMS\cite{kumar2021groomed}} & - & 19.67 & 14.32 & 11.27\\
        \multirow{1}{*}{Ground-Aware\cite{liu2021ground}} & - & 23.63 & 16.16 & 12.06\\
        \multirow{1}{*}{GUP Net\cite{lu2021geometry}} & - & 22.76 & 16.46 & 13.72\\
        \multirow{1}{*}{DEVIANT\cite{kumar2022deviant}} & - & 24.63 & 16.54 & 14.52\\
        \hline
        \multirow{1}{*}{\textbf{OriCon3D (ours)}} & - & \textbf{30.82} & \textbf{25.39} & \textbf{23.45}\\
         \hline        
    \end{tabular}
    \label{table:kitti_val}
\end{table}

\begin{table}[ht!b]
    \centering
    \caption{ Results on KITTI Test Cars at $IoU_{3D}$ $\geq$0.7}
    \begin{tabular}{ |M{3.4cm}|M{0.8cm}|M{0.8cm}M{0.8cm}M{0.8cm}|}
        \hline
        \multicolumn{1}{|c|}{} & \multicolumn{1}{c|}{Extra} & \multicolumn{3}{c|}{$AP_{3D|R_{40}}$(\%)}\\
        Method & training & Easy & Moderate & Hard\\
        \hline
        \multirow{1}{*}{PCT\cite{wang2021progressive}} & Depth & 21.00 & 13.37 & 11.31\\
        \multirow{1}{*}{DFR-Net\cite{zou2021devil}} & Depth & 19.40 & 13.63 & 10.35\\
        \multirow{1}{*}{MonoDistill\cite{chong2022monodistill}} & Depth & 22.97 & 16.03 & 13.60\\
        \multirow{1}{*}{CaDDN\cite{reading2021categorical}} & LiDAR & 19.17 & 13.41 & 11.46\\
        \multirow{1}{*}{PatchNet-C\cite{simonelli2021we}} & LiDAR & 22.40 & 12.53 & 10.60\\
        \multirow{1}{*}{DD3D\cite{park2021pseudo}} & LiDAR & 23.22 & 16.34 & 14.20\\
        \multirow{1}{*}{Kinematic\cite{brazil2020kinematic}} & Video & 19.76 & 14.10 & 10.47\\
        \hline
        \multirow{1}{*}{MonoRCNN\cite{shi2021geometry}} & - & 18.36 & 12.65 & 10.03\\
        \multirow{1}{*}{MonoDIS-M\cite{simonelli2020disentangling}} & - & 16.54 & 12.97 & 11.04\\
        \multirow{1}{*}{GrooMeD-NMS\cite{kumar2021groomed}} & - & 18.10 & 12.32 & 9.65\\
        \multirow{1}{*}{Ground-Aware\cite{liu2021ground}} & - & 21.65 & 13.25 & 9.91\\
        \multirow{1}{*}{GUP Net\cite{lu2021geometry}} & - & 20.11 & 14.20 & 11.77\\
        \multirow{1}{*}{MonoFlex\cite{zhang2021objects}} & - & 19.94 & 13.89 & 12.07\\
        \multirow{1}{*}{DEVIANT\cite{kumar2022deviant}} & - & 21.88 & 14.46 & 11.89\\
        \multirow{1}{*}{MonoCon\cite{liu2022learning}} & - & 22.50 & 16.46 & 13.95\\
        \multirow{1}{*}{CMKD\cite{hong2022cross}} & - & 28.55 & 18.69 & 16.77\\
        \multirow{1}{*}{CIE\cite{ye2022consistency}} & - & \textbf{31.55} & 20.95 & 17.83\\
        \hline
        \multirow{1}{*}{\textbf{OriCon3D (ours)}} & - & 29.66 & \textbf{23.09} & \textbf{19.33}\\
         \hline        
    \end{tabular}
    \label{table:kitti_test_car}
\end{table}

\begin{table*}[ht!b]
    \centering
    \caption{ Results on KITTI Test cyclists and pedestrians (Cyc/Ped) at $IoU_{3D}$ $\geq$0.5}
    \begin{tabular}{ |M{3.4cm}|M{0.8cm}|M{0.8cm}M{0.8cm}M{0.8cm}|M{0.8cm}M{0.8cm}M{0.8cm}|}
        \hline
        \multicolumn{1}{|c|}{} & \multicolumn{1}{c|}{Extra} & \multicolumn{3}{c|}{$Cyc AP_{3D|R_{40}}$} & \multicolumn{3}{c|}{$Ped AP_{3D|R_{40}}$}\\
        Method & training & Easy & Moderate & Hard & Easy & Moderate & Hard\\
        \hline
        \multirow{1}{*}{DDMP-3D\cite{wang2021depth}} & Depth & 4.18 & 2.50 & 2.32 & 4.93 & 3.55 & 3.01\\
        \multirow{1}{*}{DFR-Net\cite{zou2021devil}} & Depth & 5.69 & 3.58 & 3.10 & 6.09 & 3.62 & 3.39\\
        \multirow{1}{*}{MonoDistill\cite{chong2022monodistill}} & Depth & 5.53 & 2.81 & 2.40 & 12.79 & 8.17 & 7.45\\
        \multirow{1}{*}{CaDDN\cite{reading2021categorical}} & LiDAR & 7.00 & 3.41 & 3.30 & 12.87 & 8.14 & 6.76\\
        \multirow{1}{*}{DD3D\cite{park2021pseudo}} & LiDAR & 2.39 & 1.52 & 1.31 & 13.91 & 9.30 & 8.05\\
        \hline
        \multirow{1}{*}{MonoDIS-M\cite{simonelli2020disentangling}} & - & 1.17 & 0.54 & 0.48 & 7.79 & 5.14 & 4.42\\
        \multirow{1}{*}{GUP Net\cite{lu2021geometry}} & - & 4.18 & 2.65 & 2.09 & 14.72 & 9.53 & 7.87\\
        \multirow{1}{*}{MonoFlex\cite{zhang2021objects}} & - & 3.39 & 2.10 & 1.67 & 11.89 & 8.16 & 6.81\\
        \multirow{1}{*}{DEVIANT\cite{kumar2022deviant}} & - & 5.05 & 3.13 & 2.59 & 13.43 & 8.65 & 7.69\\
        \multirow{1}{*}{CMKD\cite{hong2022cross}} & - & 9.60 & 5.24 & 4.50 & \textbf{17.79} & \textbf{11.69} & \textbf{10.09}\\
        \hline
        \multirow{1}{*}{\textbf{OriCon3D (ours)}} & - & \textbf{13.47} & \textbf{8.66} & \textbf{6.25} & 16.97 & 11.11 & 9.23\\
         \hline        
    \end{tabular}
    \label{table:kitti_test_cy_ped}
\end{table*}


\section{Results}

Our OriCon3D model underwent training on an A-2000 NVIDIA GPU, employing specific hyperparameter initializations that spanned 25 epochs (equivalent to 23,438 iterations). The training configuration included a momentum value of 0.99, with weighted orientation and confidence variables set at 0.7 and 0.6, respectively. The training utilized a batch size of 8 and a learning rate of 0.0001. Notably, the training process involved batch processing with the participation of 6 workers for GPU data loading. For assessing the impact of various backbone networks on the OriCon3D architecture, we employ four metrics: AOS (Average Orientation Estimation), AP (Average Precision over 40 recall points), OS (Orientation Score), and IOU 3D (Intersection-over-union in 3D space), spanning three categories—cars, cyclist, and pedestrian. The outcomes are detailed in Table \ref{table:backbone}, where backbone architectures powered by EfficientNet and MobileNet consistently outperform the VGG-19 counterpart across all metrics and categories on the KITTI Val dataset. While comparing EfficientNet and MobileNet-powered architectures reveals mixed results, with EfficientNet excelling in some metrics and MobileNet in others, the former consistently achieves higher scores on the AP metric across all categories. Fig. \ref{fig:qualitative} displays qualitative results on the KITTI dataset. Each column, in top-down order, presents the 2D bounding box prediction, 3D bounding box ground truth, and predicted 3D bounding boxes with VGG-19, MobileNet-v2, and EfficientNet-v2 backbones. Given that the AP metric is pivotal for gauging algorithm performance on the official KITTI Monocular 3D object detection benchmark, we leverage our OriCon3D architecture with the EfficientNet backbone network for comparative assessments against other state-of-the-art architectures. Examining Table \ref{table:kitti_val}, our architecture, exclusively trained on images, surpasses all other architectures on the KITTI validation dataset\cite{chen20153d} for the 'car' category, except for \cite{wang2021progressive}, which benefits from an extra depth modality. Furthermore, in evaluations on the Official KITTI dataset for the 'car' category, our architecture outperforms all other state-of-the-art architectures across moderate and hard difficulty levels, as illustrated in Table \ref{table:kitti_test_car}. Notably, the OriCon3D architecture demonstrates superior performance for the 'cyclist' category and remains competitive for the 'pedestrian' category, as indicated in Table \ref{table:kitti_test_cy_ped}.

\section{Conclusion}
In this work, we identify the low variance of the bounding box dimensions across any category for 3D object detection. Based on this insight, we propose a novel two-stage architecture, OriCon3D, multibin architecture that directly regresses over the orientation angle, and confidence. With the leverage of varying backbone networks, we observe consistent outperformance of EfficientNet-v2 and MobileNet-v2 over VGG-19 across key metrics and categories, on a standard KITTI validation dataset. Evaluated on the Official KITTI test dataset, OriCon3D with EfficientNet achieves top-tier performance, surpassing competitors for 'car' and 'cyclist' categories, and maintaining competitiveness for 'pedestrian.' These results highlight OriCon3D's robustness, showcasing its potential in real-world applications for autonomous systems.

\bibliographystyle{IEEEtran}
\bibliography{IEEEabrv,reference.bib}

\end{document}